# Cells are Actors: Social Network Analysis with Classical ML for SOTA Histology Image Classification

Neda Zamanitajeddin[1], Mostafa Jahanifar[1], and Nasir Rajpoot[1]

[1] Tissue Image Analytics Centre, Department of Computer Science, University of Warwick, Coventry, UK
`neda.zamanitajeddin@warwick.ac.uk`

**Abstract.** Digitization of histology images and the advent of new computational methods, like deep learning, have helped the automatic grading of colorectal adenocarcinoma cancer (CRA). Present automated CRA grading methods, however, usually use tiny image patches and thus fail to integrate the entire tissue micro-architecture for grading purposes. To tackle these challenges, we propose to use a statistical network analysis method to describe the complex structure of the tissue micro-environment by modelling nuclei and their connections as a network. We show that by analyzing only the interactions between the cells in a network, we can extract highly discriminative statistical features for CRA grading. Unlike other deep learning or convolutional graph-based approaches, our method is highly scalable (can be used for cell networks consist of millions of nodes), completely explainable, and computationally inexpensive. We create cell networks on a broad CRC histology image dataset, experiment with our method, and report state-of-the-art performance for the prediction of three-class CRA grading.

## 1      Introduction

With the advances of digital pathology and the advent of sophisticated computerized methods (like deep learning), achieving reliable computer-assisted diagnostic systems for pathology applications has become conceivable. Many automatic approaches have been introduced in the literature to work with large histology images also known as the Whole Slide Images (or WSIs) for automatic cancer diagnosis or patient prognosis [1-5], most of which rely on convolutional neural networks (CNN) [6]. Most existing methods are unable to handle large WSI and often follow a patch prediction paradigm, where the WSI is divided into small patches for CNN based prediction and finally all predictions are aggregated to achieve the final decision. In this paradigm, critical contextual information from the larger field of view (FOV) in WSIs can be lost. Recent works on context-aware CNNs have aimed at addressing that problem by processing multi-scale fields of view (FOVs) [7, 8]. However, those methods are still bounded by the GPU memory limitation. Lack of interpretability is another limitation faced by deep CNN features. In particular, the connection of deep CNN features with tissue morphology or structure of pathology primitives in a histology image is unknown. To avoid these shortcomings, methods based on graph theory have been proposed in the area of computational pathology where the histology primitives (nuclei, glands, etc.) and their interactions are modelled as a graph [9-12]. For example, Javed *et al*. [11, 13] showed that constructing graphs of histology image patches and finding cell communities in the graphs can be used to classify tissue phenotypes. However, in these works, it is unclear how to choose and integrate graph-level features to better reflect the diverse arrangement of cells in various tissue components. Zhou *et al*. [14] proposed a sophisticated Graph Convolutional Network (GCN) based approach, called CGC-Net, for histology image classification which can automatically extract features from histology graphs, classify them, and achieve state-of-the-art performance. Although CGC-Net can accept larger contextual patches than CNNs, it is a computationally expensive method, the GCN features lack interpretability and are still restricted by the GPU memory size.

In this paper, we propose to tackle these problems and approach the classification of histology images using a novel social network analysis (SNA) paradigm [15]. Cell graphs from WSIs may contain millions of nodes, edges (connections) and cellular communities which contain diagnostically relevant information about tissue micro-environment that may be difficult to quantify by visual inspection. In this work, we incorporate SNA measures to translate the raw information of cell-to-cell connections in cell graphs into perceivable and interpretable features that can help distinguishing tissue phenotypes or highlight the biological significance of tissue regions. We propose three different approaches to utilize SNA measures as features in a classical machine learning setting for automatic image-level Colorectal Adenocarcinoma (CRA) grading from patch-level cell graphs, achieving state-of-the-art performance. Our approach does not rely on precise nuclei segmentation or feature extraction as it extracts information from the cell network by investigating only cell-cell connections and encoding spatial relationships between cells. It is computationally efficient and is not bounded by GPU memory, which makes it easily scalable.

To the best of our knowledge, this is the first study that proposes the analysis of histology images using SNA measures for automatic CRA grading, demonstrating the role of cells and cellular communities as actors. The rest of the paper is organized as follows. In section 2, we describe details of the proposed methodology, afterwards, results of applying our method on a CRA grading dataset are reported and discussed in section 3. Finally, we conclude the paper with a summary of our findings and future directions.

## 2       Materials and Methods

In this section, we will describe our proposed method of using social network analysis techniques for the classification of colorectal cancer (CRC) into three CRA degrees. First, we will describe the cell social networks construction. Then, utilized social network analysis (SNA) measures are introduced. Afterwards, statistical tools are employed to abstract the representation of SNA measures into relatively low-dimensional feature vectors. Finally, three different scenarios are proposed to aggregate patch-level information into the image-level prediction of CRA grade. A schematic overview of these steps is shown in Fig. 1. The following subsections describe these steps in more detail.

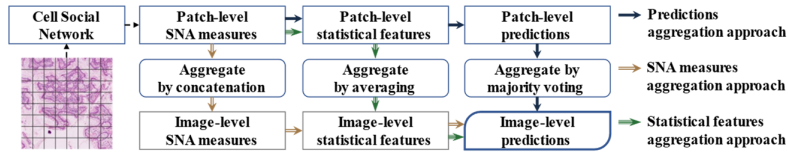

**Fig. 1.** Overview of the proposed SNA paradigm for image-level prediction.

### 2.1 Data

The proposed method is tested on the CRC dataset [16], consisting of 139 images with an average size of 4548×7520 taken from WSIs captured at 20x magnification related to 38 patients. Two experienced pathologists classified all images into normal (#71 Grade 1), low-grade (#33 Grade 2), and high-grade (#35 Grade 3) categories for CRA grading based on glandular morphology. To fairly compare the performance of the proposed SNA based features with CNN and CGC based features, like [7, 14], we split the dataset into three folds and extract large patches of 1792×1792 pixels to be used for cell network construction.

### 2.2 Cell Social Network

Our approach relies on building a meaningful network (graph) from a histology image that represents the possible cell-to-cell interactions. For such a graph $G = (V, E)$, we can consider nuclei as network nodes ($V$) and their potential interactions as network edges ($E$). Therefore, we follow these two steps to construct cell networks for each patch in the dataset: 1) node identification by nuclear instance segmentation and 2) finding the network's edge configuration.

**Node identification.** For the sake of comparison fairness, we utilize CIA-Net [17] for nuclei instance segmentation similar to Zhou et al. [14]. However, unlike [14], our approach does not need precise nuclei segmentation because we do not include nuclear appearance features in our cell network. Therefore, detecting the centroid position of each nucleus [18] would suffice. Also, there is no need for nuclei sampling because our approach is not restricted to computationally extensive CGNs.

**Edge configuration.** Hypothesizing that close cells in a histology sample are more likely to interact with each other, we consider an edge between two nuclei if their Euclidean distance is smaller than a specific radius. Also, we bound the number of edges to the *k*-nearest neighbours to avoid over-connecting a node. In other words, the adjacency matrix for nodes $i, j$ is $A_{ij} = 1$ if their Euclidean distance is smaller than a radius $D(i, j) < r$, and if $j$ belongs to the k-nearest neighbours of node $i$.

## 2.3 SNA Measures

Raw data from a large cell network is not easily understandable. Assuming that cells are social actors in the tumour micro-environment, we extract several measures from the social network analysis (SNA) discipline [15, 19], as listed below.

**Node Degree (NDe).** Node degree $\deg(v_i)$ of node $v_i$ in graph $G = (V, E)$ is simply the number of edges connected to that node. For instance, in a cell social network, a cell's degree is the number of cells it is directly connected to.

**Clustering Coefficient (CCo).** This is defined as the fraction of the possible triangles that occur across a node: $CCo_v = 2\text{Tri}(v)/(\deg(v)(\deg(v)-1))$ which quantifies the density of triangles $\text{Tri}(v)$ in a network. In a social network of cells, the clustering coefficient measures how many cells in the graph tend to cluster together.

**Closeness Centrality (CC).** In a network with $N$ nodes, the closeness centrality of a node $v_i$ is reciprocal of the average of the shortest-path distances, $d(.,.)$ between node $v_i$ and all other nodes in the network: $CC(v_i) = (N-1)/\sum_{j, j \neq i} d(v_i, v_j)$. Closeness centrality highlights nodes that can easily access other nodes. In cell social network analogy, a cell with higher closeness centrality is closer to all other cells.

**Degree Centrality (DC).** This is the simplest centrality defined for a network which is calculated by normalizing the node degree: $DC(v_i) = \sum_{j=1}^{N} A_{ij}/(N-1)$. In comparison to node degree, network size information is embedded in this measure

**Betweenness Centrality (BC).** For node $v$, $BC(v) = \sum_{s,t \in V} \sigma(s,t|v)/\sigma(s,t)$ is the betweenness centrality in which $\sigma(s,t)$ is the total number of shortest paths from node $s$ to node $t$ and $\sigma(s,t|v)$ is the number of those paths that pass through $v$. BC can highlight nodes that act as bridges connecting two parts of a network. Interpretation of this measure in cell social network depends on the clinical application, however, in some cell graphs, BC does not show significant importance [20].

**Eigen Vector Centrality (EVC).** This centrality measures the influence of a node in the network. EVC is an extension of degree centrality in which the score of each node is calculated based on the scores of its neighbouring nodes. Therefore, a node in the network can have a high EVC score if it is connected to either many nodes or some nodes with high degrees. The eigenvector centrality for node $i$ is $x_i$, $x_i = \lambda^{-1} \sum_j A_{ij} x_j$ where $A$ is network adjacency matrix with eigenvalue of $\lambda$ [15]. EVC in cell social network highlights the cells which hold wide-reaching influence in the network.

**Katz Centrality (KC).** Similar to EVC, Katz centrality is a generalization of degree centrality where all connected neighbouring nodes, immediate or distant, are taken into consideration when calculating each node's centrality score [21]. In KC every neighbouring node will be assigned an initial constant centrality $\beta$, and contributions of distant nodes are penalized with attenuation factor $\alpha$, as in $x_i = \alpha \sum_j A_{ij} x_j + \beta_i$.

### 2.4 SNA-based Statistical Features

We extract statistical features from each SNA measure of a cell network to represent that network with a relatively low-dimensional fixed-size feature vector. For each cell social network, we find the distribution of values of every SNA measure by calculating their histogram and then concatenate the resulting histogram counts and bins alongside with maximum, mean, and standard deviation of those values to build the feature vector. In total, we extract 180 features from all SNA measures for each cell network.

### 2.5 Generating image-level prediction

We propose three approaches to aggregate patch-level information for image-level predictions. A schematic overview of these three approaches is shown in Fig. 1.
**Predictions aggregation.** We train an SVM [22] on statistical feature vectors extracted from patch-level cell networks. Having patch-level predictions, majority voting is utilized to aggregate them into the final image-level label.
**Network measure aggregation.** In this approach, we aggregate SNA measures for all patches of an image by simply concatenating them to form the image-level SNA measures. Having these, we extract statistical features for each image as described in section 2.4 and they can be used to train an SVM model for image-level classification.
**Statistical feature aggregation.** Here, we average patch-level statistical features of the patch-level SNA measures over patches belonging to each image.

To avoid overfitting and choosing the optimal subset of features, we incorporated sequential features selection (SFS) [23]. In all experiments, we used an SVM classifier [22] with an RBF kernel. Also, SVM training and evaluation are done in a 3-fold cross-validation framework similar to [7, 14] for a fair comparison.

## 3    Results and Discussions

### 3.1 Feature Selection for CRA Grade Prediction

In this section, we report the image-level accuracy of classifying the CRC dataset [16] in three CRA grades. In Table 1, the results of cross-validation experiments using different methods are reported in which the last three rows correspond to proposed methods in the current work. Utilizing social network analysis techniques, we achieve an image-level accuracy of 96.27%, 97.8%, and 99.24% using patch-level predictions, network measures, and statistical features aggregation approaches, respectively, outperforming all other approaches in the literature by a large margin. Keep in mind that these values are achieved using only a handful of selected features from sequential feature selection experiments as illustrated in Fig. 2 for different aggregation scenarios. Furthermore, we observed that using a Random Forest classifier instead of SVM in the proposed framework would cause a decrease in performance, achieving the accuracy of 95.2%, 95.5%, 96.3% for patch prediction aggregation, SNA measures aggregation, and statistical features aggregation scenarios, respectively.

Table 1. Performance comparison of different CRA grading methods.

| Method | Acc % |
|---|---|
| BAM-entropy | 90.66±2.45 |
| Context-G | 89.96±3.54 |
| ResNet50 | 92.08±2.08 |
| CA-CNN | 95.70±3.04 |
| CGC-Net | 97.00±1.10 |
| Predictions agg. | 96.27±2.82 |
| SNA measures agg. | 97.80±2.17 |
| **Statistical features agg.** | **99.24±1.31** |

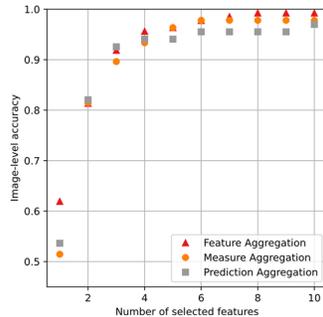

**Fig. 2.** Feature selection experiments for different aggregation approaches.

Some of the top-performing selected features are the number of nodes whose clustering coefficient attribute is in the range (0.4, 0.5) and range (0.9, 1), the number of nodes with node degree of 10 and 12, the number of nodes that have eigenvector centrality attribute less than 0.005, and the number of nodes that their closeness centrality attribute is less than 0.001. Our feature selection experiments show that there are always statistical features derived from clustering coefficient measures in the selected set in all scenarios whereas none of them has features related to Betweenness centrality. Having said that, discussing specific ranges from the distribution of SNA measures can be more intuitive and explain the biological significance of these measures (refer to Sections 3.2 and 3.3).

In comparison with the baseline method, BAM [16], which is based on glandular morphology analysis, our best performing method gains over 8.52% accuracy. By comparing the results of predictions aggregation scenario with other methods that similarly use majority voting over patch-level predictions (Context-G [8], C A-CNN [7], and ResNet50 [24]), one can infer that patch-level statistical features extracted from SNA measures are very discriminative for CRA grading. On the other hand, using statistical feature aggregation can lead to better image-level performance (2.24% better than CGC-Net in accuracy). Moreover, our method can be easily scaled up to work directly on WSI-based cell networks, while methods like CGC-Net [14], CA-CNN [7] or any other deep learning-based models cannot abide by such tasks due to hardware limitation. This shows the effectiveness and efficiency of the proposed SNA based method.

We trained and tested the patch classifier model (in the first aggregation scenario) based on the labels of the large images whereas there are some regions (patches) in Grade 2, 3 samples that look normal and should be labelled as Grade 1. This noise in patch prediction refrains the model to learn optimal weights for image-level prediction. On the other hand, image-level classifiers in the other two aggregation methods are using information over the whole image area to predict a single label per image. The aggregation of information obviate the mentioned noise effect and leads to better results. Also, Table 1 suggests that local statistics of SNA measures are more important than statistics of global aggregation of SNA measures, as the statistical feature aggregation method performs better.

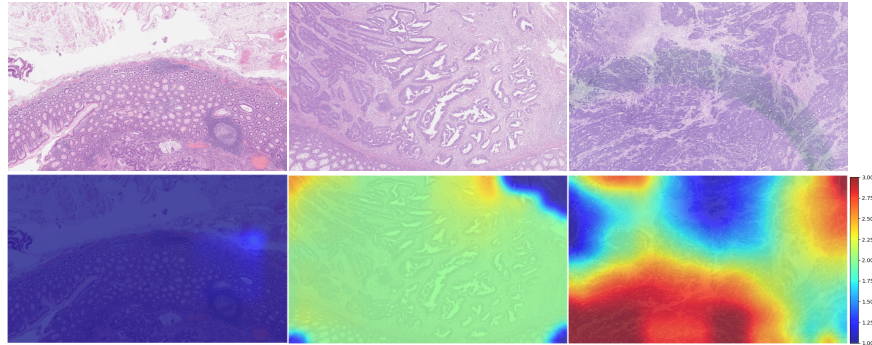

**Fig. 3.** Image-level prediction heatmap overlayed on images of three CRA grades (left to right: Grades 1, 2, and 3). The color bar indicates the estimated CRA grade score for regions.

The qualitative performance of the selected statistical features for identifying different tissue phenotypes in large histology images is also illustrated in Fig. 3. These saliency maps are generated by predicting labels for overlapping image patches. Although selected features are chosen based on image-level classification performance, samples in Fig. 3 show that these features can perform very well on patch-level classification and meaningfully detect different tissue phenotypes present in the image.

### 3.2 Distributions of SNA Measures

To better represent the discriminative power of the proposed SNA-based statistical features, we investigate the difference in average distributions of various SNA measures over each CRA grade using the student's $t$-test. For each SNA measure, the $t$-test is done in a pairwise manner between three different grades (Grade 1 vs. Grade 2, Grade 1 vs. Grade 3, and Grade 2 vs. Grade 3) and results are summarized in Table 2. Values in Table 2 indicate that pairwise differences in distributions of most SNA measures over three grades are significantly different (*p*-value<0.05), except for Betweenness Centrality (BC). This observation acknowledges that extracted statistical features from SNA measures are powerful enough for CRA grade discrimination. This is also visually verified by plotting the average distributions for "Katz Centrality", "Clustering Coefficient" and "Closeness Centrality" SNA measures in Fig. 4.

**Table 2.** Results of statistical comparison ($t$-test) experiments on SNA measure distributions.

| CRA types | *p*-value | | | | | | |
|---|---|---|---|---|---|---|---|
| | *ND* | *CCo* | *DC* | *BC* | *CC* | *EVC* | *KC* |
| Grade 1 vs. Grade 2 | **0.01** | **0.003** | **0.0007** | 0.1 | **<1e-23** | **<7e-5** | **<4e-26** |
| Grade 1 vs. Grade 3 | 0.05 | **0.006** | **<8e-5** | 0.2 | **<5e-9** | **0.0003** | **1e-25** |
| Grade 2 vs. Grade 3 | 0.2 | **0.04** | **0.04** | 0.3 | **0.001** | **0.01** | **<8e-11** |

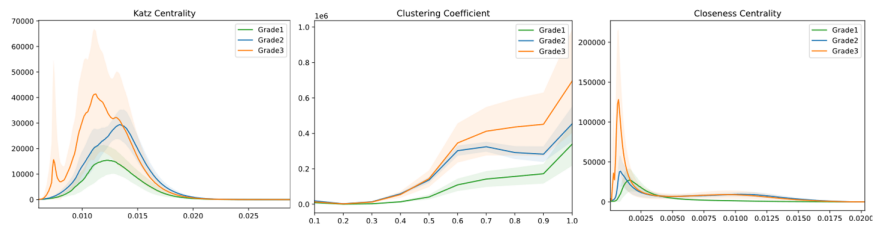

**Fig. 4.** Average distribution of three SNA measures from CRA samples in different grades.

### 3.3 Understanding the biological significance of SNA measures

As mentioned above, high values of node clustering coefficient indicate that there is a more significant intra-cell interaction present in the network and nodes tend to create tightly knit groups which are usually due to hyper aggregation of nuclei in tumour regions. Therefore, the high number of nodes with a high value of clustering coefficient can indicate the presence of a high-grade CRA community in the tissue. This phenomenon is easily observable in Fig. 4 and the fourth row of Fig. 5, where nodes coloured based on the values of the SNA measure represented in that row.

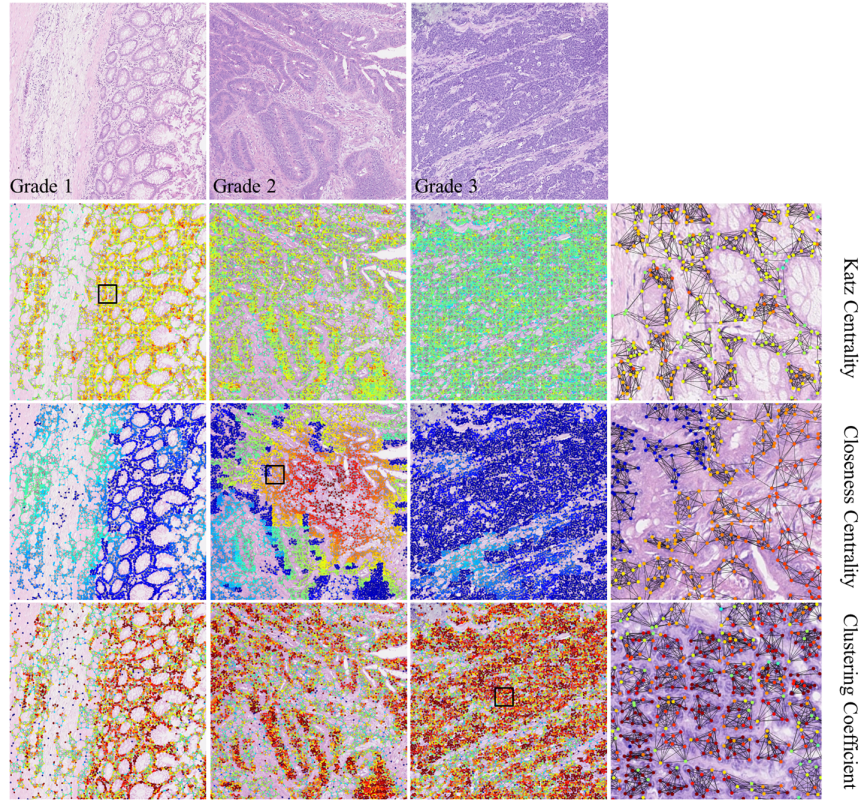

**Fig. 5.** Sample images from three CRA grades and their corresponding cell graphs overlaid on the original images. Color of each node is based on the values of 3 specific SNA measures with their minimum and maximum values shown in blue and red, respectively. Final column shows zoomed-in regions (shown as black squares) in one of the images for each of the grades.

Closeness centrality operates the exact opposite, where cells that are in a close-knit and dense community would be assigned smaller values because the average short-path distance increases throughout the network with the increased number of cells. Therefore, in Grade 3 phenotypes we would have a large number of cells with low closeness centrality and we expect that distribution of this SNA measure for high-grade cases be higher around small closeness values, as depicted in Fig. 4 and the third row and last column of Fig. 5. Stroma and non-mucosa regions are visibly differentiable in Grade 1 samples, based on closeness centrality in Fig. 5 while in Grade 3, the ***tumour-stromal interface*** seems segmentable based on this measure. Also, It is obvious in the second row of Fig. 5 that Katz centrality can assign values to every node in the network by considering the degrees of all nodes, resulting in a visible difference in node measure values for different CRA grades. These examples show the power of SNA measures in highlighting the biological significance of every node in the cell network.

## 4    Conclusion

In this paper, we proposed to use social network analysis for histology image classification which relies only on the locations and interactions of cells. We showed how SNA measures can highlight the biological importance of cells and tissue regions in the histology image while their statistics significantly segregate between different types of CRA grades. We were able to achieve SOTA accuracy (99.24%) in CRA grading using only a handful of statistical features extracted from SNA measures. Being highly performant and non-reliant on GPUs (intensive computation), we believe that SNA based approaches hold promise for analyzing large cell networks and deserve further investigation for other cancer diagnosis or prognosis applications.

108. Sirinukunwattana, K., et al: Improving whole slide segmentation through visual context-a systematic study. In: International Conference on Medical Image Computing and Computer-Assisted Intervention. Springer (2018).
9. Bilgin, C.C., et al.: ECM-aware cell-graph mining for bone tissue modeling and classification. Data mining and knowledge discovery 20(3), 416-438 (2010).
10. Demir, C., S.H. Gultekin, and B. Yener: Augmented cell-graphs for automated cancer diagnosis. Bioinformatics 21(suppl_2), ii7-ii12 (2005).
11. Javed, S., et al.: Cellular community detection for tissue phenotyping in colorectal cancer histology images. Medical image analysis 63, 101696 (2020).
12. Sirinukunwattana, K., et al.: Novel digital signatures of tissue phenotypes for predicting distant metastasis in colorectal cancer. Scientific Reports 8(1), 1-13 (2018).
13. Javed, S., et al.: Multiplex Cellular Communities in Multi-Gigapixel Colorectal Cancer Histology Images for Tissue Phenotyping. IEEE Transactions on Image Processing 29, 9204-9219 (2020).
14. Zhou, Y., et al.: Cgc-net: Cell graph convolutional network for grading of colorectal cancer histology images. In: Proceedings of the IEEE/CVF International Conference on Computer Vision Workshops. (2019).
15. Newman, M.: Networks. Oxford university press (2018).
16. Awan, R., et al.: Glandular morphometrics for objective grading of colorectal adenocarcinoma histology images. Scientific Reports 7(1), 1-12 (2017).
17. Zhou, Y., et al.: Cia-net: Robust nuclei instance segmentation with contour-aware information aggregation. In: International Conference on Information Processing in Medical Imaging. Springer (2019).
18. Raza, S.E.A., et al.: Deconvolving convolutional neural network for cell detection. In: 2019 IEEE 16th International Symposium on Biomedical Imaging (ISBI 2019). IEEE (2019).
19. Scott, J.: Social network analysis. Sociology 22(1), 109-127 (1988).
20. Failmezger, H., et al.: Topological Tumor Graphs: a graph-based spatial model to infer stromal recruitment for immunosuppression in melanoma histology. Cancer research 80(5), 1199-1209 (2020).
21. Katz, L.: A new status index derived from sociometric analysis. Psychometrika 18(1), 39-43 (1953).
22. Bishop, C.M.: Pattern recognition and machine learning. Springer (2006).
23. Jahanifar, M., M. Hasani, and S.J. Khaleghi: Automatic zone identification in blood smear images using optimal set of features. In 2016 23rd Iranian Conference on Biomedical Engineering and 2016 1st International Iranian Conference on Biomedical Engineering (ICBME). IEEE (2016).
24. He, K., et al.: Deep residual learning for image recognition. In Proceedings of the IEEE conference on computer vision and pattern recognition. (2016).